\documentclass{article} 
\usepackage{general,times}

\usepackage{amsmath,amsfonts,bm}

\def\eqref#1{equation~\ref{#1}}

\def\1{\bm{1}}

\DeclareMathAlphabet{\mathsfit}{\encodingdefault}{\sfdefault}{m}{sl}
\SetMathAlphabet{\mathsfit}{bold}{\encodingdefault}{\sfdefault}{bx}{n}

\newcommand*{\QEDW}{\hfill\ensuremath{\square}}%
\newcommand{\norm}[1]{\left\lVert#1\right\rVert}

\newcommand{\br}[1]{\left\{#1\right\}}

\newcommand{\dist}{\mathrm{dist}}

\newcommand{\comment}[1]{}
\newcommand{\REAL}{\ensuremath{\mathbb{R}}}

\newif\ifproofs
\proofsfalse

\usepackage{enumitem}
\usepackage{hyperref}
\usepackage{url}
\usepackage{graphicx} 
\usepackage{natbib}  

\title{Deep Learning Meets Projective Clustering}

\author{Alaa Maalouf$^1$\thanks{equal contribution}~ , Harry Lang$^2$\footnotemark[1]~ , Daniela Rus$^2$ \& Dan Feldman$^1$ \\
$^1$ Robotics \& Big Data Labs, Department of Computer Science,
University of Haifa\\
$^2$ CSAIL, MIT\\
\texttt{alaamalouf12@gmail.com, harry1@mit.edu,} \\ \texttt{rus@csail.mit.edu, dannyf.post@gmail.com}
}

\begin{document}

\maketitle

\begin{abstract}
A common approach for compressing NLP networks is to encode the embedding layer as a matrix $A\in\mathbb{R}^{n\times d}$, compute its rank-$j$ approximation $A_j$ via SVD, and then factor $A_j$ into a pair of matrices that correspond to smaller fully-connected layers to replace the original embedding layer. Geometrically, the rows of $A$ represent points in $\mathbb{R}^d$, and the rows of $A_j$ represent their projections onto the $j$-dimensional subspace that minimizes the sum of squared distances (``errors'') to the points. 
In practice, these rows of $A$ may be spread around $k>1$ subspaces, so factoring $A$ based on a single subspace may lead to large errors that turn into large drops in accuracy.

Inspired by \emph{projective clustering} from computational geometry,  we suggest replacing this subspace by a set of $k$ subspaces, each of dimension $j$, that minimizes the sum of squared distances over every point (row in $A$) to its \emph{closest} subspace. Based on this approach, we provide a novel architecture that replaces the original embedding layer by a set of $k$ small layers that operate in parallel and are then recombined with a single fully-connected layer. 

Extensive experimental results on the GLUE benchmark yield networks that are both more accurate and smaller compared to the standard matrix factorization (SVD). For example, we further compress DistilBERT by reducing the size of the embedding layer by $40\%$ while incurring only a $0.5\%$ average drop in accuracy over all nine GLUE tasks, compared to a $2.8\%$ drop using the existing SVD approach.
On RoBERTa we achieve $43\%$ compression of the embedding layer with less than a $0.8\%$ average drop in accuracy as compared to a $3\%$ drop previously.
Open code for reproducing and extending our results is provided.
\end{abstract}

\section{Introduction and Motivation}\label{sec:matrixfactorization}
Deep Learning revolutionized Machine Learning by improving the accuracy by dozens of percents for fundamental tasks in Natural Language Processing (NLP) through learning representations of a natural language via a deep neural network~\citep{mikolov2013distributed,radford2018improving,le2014distributed,peters2018deep,radford2019language}. Lately, it was shown that there is no need to train those networks from scratch each time we receive a new task/data, but to fine-tune a full pre-trained model on the specific task~\citep{dai2015semi,radford2018improving,devlin-etal-2019-bert}.
However, in many cases, those networks are extremely large compared to classical machine learning models. 
For example, both BERT~\citep{devlin-etal-2019-bert} and XLNet~\citep{yang2019XLNet} have more than $110$ million parameters, and RoBERTa~\citep{liu2019RoBERTa} consists of more than $125$ million parameters.
Such large networks have two main drawbacks: (i) they use too much storage, e.g. memory or disk space, which may be infeasible for small IoT devices, smartphones, or when a personalized network is needed for each user/object/task, and (ii) classification may take too much time, especially for real-time applications such as NLP tasks: speech recognition, translation or speech-to-text.

\begin{figure}[h]
    \includegraphics[width=\textwidth]{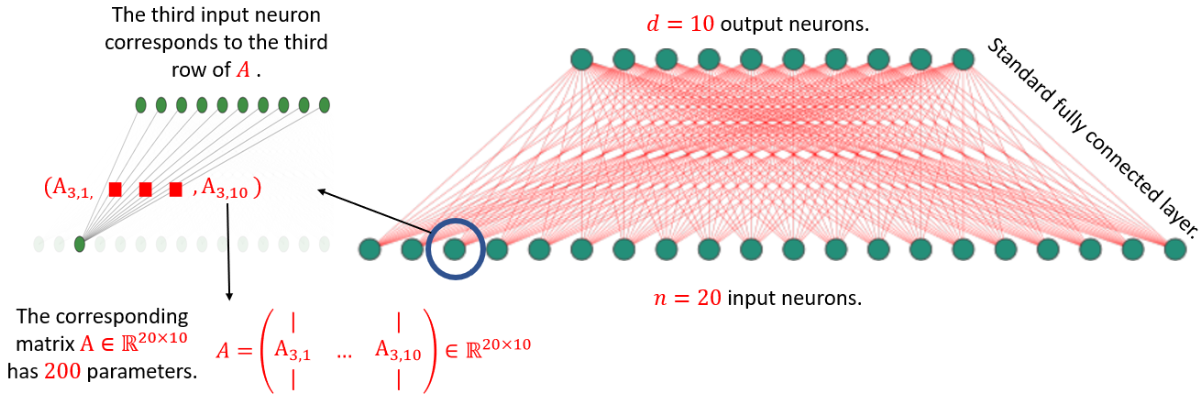}
    \caption{A standard embedding (or fully-connected) layer of $20$ input neurons and $10$ output neurons. Its corresponding matrix $A\in \REAL^{20\times 10}$ has $200$ parameters, where the $i$th row in $A$ is the vector of weights of the $i$ neuron in the input layer. }
    \label{fig:fc}
\end{figure}

\paragraph{Compressed Networks. }To this end, many papers suggested different techniques to compress large NLP networks, e.g., by low-rank factorization~\citep{wang2019structured,lan2019albert}, pruning~\citep{mccarley2019pruning,michel2019sixteen,fan2019reducing,guo2019reweighted,gordon2020compressing}, quantization~\citep{zafrirq8bert,shen2020q}, weight sharing~\citep{lan2019albert}, and knowledge distillation~\citep{sanhdistilbert,tang2019distilling,mukherjee2019distilling,liu2019attentive,sun2019patient,jiao2019tinybert}; see more example papers and a comparison table in~\citet{gordon_2019} for compressing the BERT model. There is no consensus on which approach should be used in what contexts. However, in the context of compressing the embedding layer, the most common approach is low-rank factorization as in~\citet{lan2019albert}, and it may be combined with other techniques such as quantization and pruning.

In this work, we suggest a novel low-rank factorization technique for compressing the embedding layer of a given model. This is motivated by the fact that in many networks, the embedding layer accounts for $20\%-40\%$ of the network size. Our approach - MESSI: Multiple (parallel) Estimated SVDs for Smaller Intralayers - achieves a better accuracy for the same compression rate compared to the known standard matrix factorization. To present it, we first describe an embedding layer, the known technique for compressing it, and the geometric assumptions underlying this technique. Then, we give our approach followed by geometric intuition, and detailed explanation about the motivation and the architecture changes. Finally, we report our experimental results that demonstrate the strong performance of our technique. 

\paragraph{Embedding Layer. }The embedding layer aims to represent each word from a vocabulary by a real-valued vector that reflects the word's semantic and syntactic information that can be extracted from the language. 
One can think of the embedding layer as a simple matrix multiplication as follows. The layer receives a standard vector $x\in \REAL^n$ (a row of the identity matrix, exactly one non-zero entry, usually called \emph{one-hot vector}) that represents a word in the vocabulary, it multiplies $x$ by a matrix $A^T\in \REAL^{d\times n}$ to obtain the corresponding $d$-dimensional word embedding vector $y=A^Tx$, which is the row in $A$ that corresponds to the non-zero entry of $x$. 
The embedding layer has $n$ input neurons, and the output has $d$ neurons. The $nd$ edges between the input and output neurons define the matrix $A\in\REAL^{n\times d}$. Here, the entry in the $i$th row and $j$th column of $A$ is the weight of the edge between the $i$th input neuron to the $j$th output neuron; see Figure.~\ref{fig:fc}. 

\paragraph{Compressing by Matrix Factorization. }A common approach for compressing an embedding layer is to compute the $j$-rank approximation $A_j\in \REAL^{n\times d}$ of the corresponding matrix $A$ via SVD (Singular Value Decomposition; see e.g.,~\citet{lan2019albert,yu2017compressing} and~\citet{acharya2019online}), factor $A_j$ into two smaller matrices $U\in \REAL^{n\times j}$ and $V\in \REAL^{j \times d}$ (i.e. $A_j=UV$), and replace the original embedding layer that corresponds to $A$ by a pair of layers that correspond to $U$ and $V$.
The number of parameters is then reduced to $j(n+d)$. Moreover, computing the output takes $O(j(n+d))$ time, compared to the $O(nd)$ time for computing $A^Tx$.
As above, we continue to use $A_j$ to refer to a rank-$j$ approximation of a matrix $A$.

\paragraph{Fine tuning.} The layers that correspond to the matrices $U$ and $V$ above are sometimes used only as initial seeds for a training process that is called \emph{fine tuning}. Here, the training data is fed into the network, and the error is measured with respect to the final classification. Hence, the structure of the data remains the same but the edges are updated in each iteration to give a better accuracy.

\begin{figure}[t]
    \includegraphics[width=\textwidth]{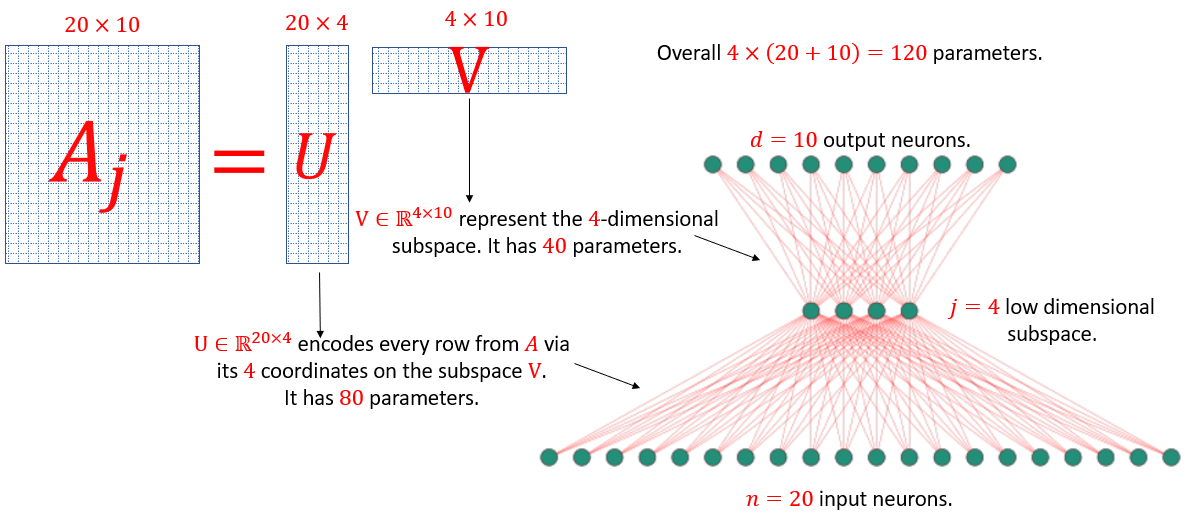}
    \caption{Factorization of the embedding layer (matrix) $A\in \REAL^{20\times 10}$ from Figure~\ref{fig:fc} via standard matrix factorization (SVD) to obtain two smaller layers (matrices) $U\in \REAL^{20\times 4}$ and $V\in \REAL^{4\times 10}$. In this example, the factorization was done based on a $4$-dimensional subspace. The result is a compressed layer that consists of $120$ parameters. The original matrix had $200$ parameters. See more details in the figure.}
    \label{fig:svd}
\end{figure}

\paragraph{Geometric intuition. }The embedding layer can be encoded into a matrix $A\in\REAL^{n\times d}$ as explained above. Hence, each of the $n$ rows of $A$ corresponds to a point (vector) in $\REAL^d$, and the $j$-rank approximation $A_j\in \REAL^{n\times d}$ represents the projection on the $j$-dimensional subspace that minimizes the sum of squared distances (``errors'') to the points. Projecting these points onto any $j$-dimensional subspace of $\REAL^d$ would allow us to encode every point only via its $j$-coordinates on this subspace, and store only $nj$ entries instead of the original $nd$ entries of $A$. This is the matrix $U\in \REAL^{n\times j}$, where each row encodes the corresponding row in $A$ by its $j$-coordinates on this subspace. The subspace itself can be represented by its basis of $j$ $d$-dimensional vectors ($jd$ entries), which is the column space of a matrix $V^T\in\REAL^{d\times j}$. Figure~\ref{fig:svd} illustrates the small pair of layers that corresponds to $U$ and $V$, those layers are a compression for the original big layer that corresponds to $A$. 

However, our goal is not only to compress the network or matrix, but also to approximate the original matrix operator $A$. To this end, among all the possible $j$-subspaces of $\REAL^d$, we may be interested in the $j$-subspace that minimizes the sum of squared distances to the points, i.e., the sum of squared projected errors. This subspace can be computed easily via SVD. The corresponding projections of the rows of $A$ on this subspace are the rows of the $j$-rank matrix $A_j$.

The hidden or statistical assumption in this model is that the rows of the matrix $A$ (that represents the embedding layer) were actually generated by adding i.i.d. Gaussian noise to each point in a set of $n$ points on a $j$-dimensional subspace, that is spanned by what are called latent variables or factors. Given only the resulting matrix $A$, the $j$-subspace that maximizes the likelihood (probability) of generating the original points is spanned by the $j$ largest singular vectors of $A$.

\paragraph{Why a single distribution? }Even if we accept the assumption of Gaussian noise, e.g. due to simplicity of computations or the law of large numbers, it is not intuitively clear why we should assume that the rows of $A$ were sampled from a single distribution. Natural questions that arise are:
\begin{enumerate}[label=(\roman*)]
    \item Can we get smaller and/or more accurate models in real-world networks by assuming multiple instead of a single generating distribution (i.e. multiple subspaces)?
    \item Can we efficiently compute the corresponding factorizations and represent them as part of a network ?
\end{enumerate}

\section{Our contribution}\label{sec:ourcont}
\begin{figure}[t]
    \includegraphics[scale = 0.6]{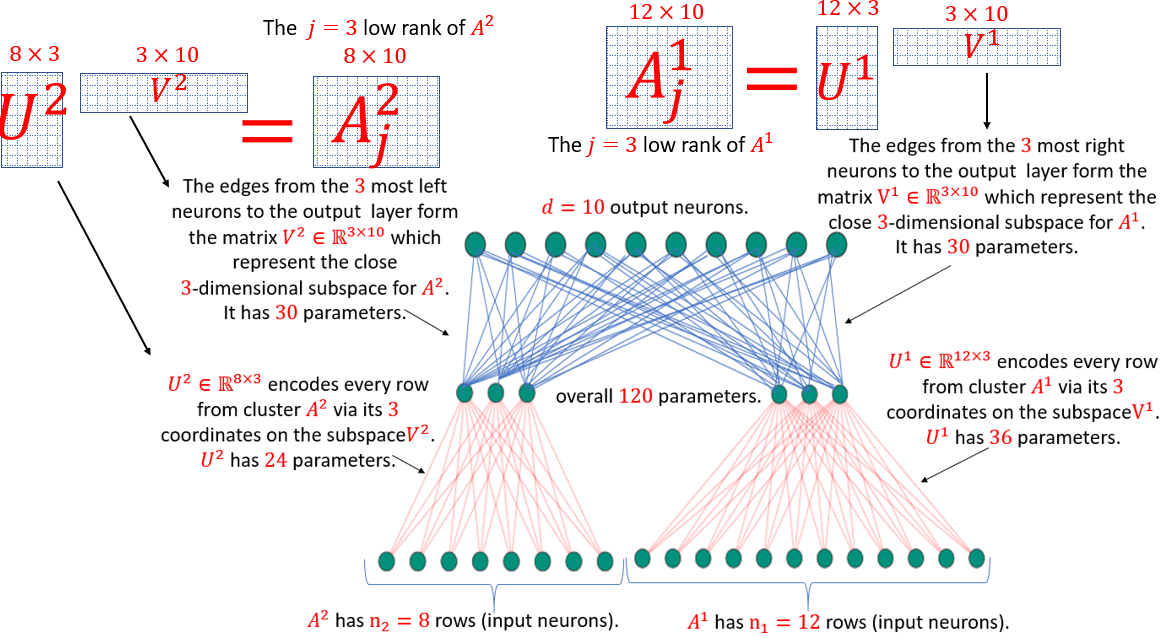}
    \caption{\textbf{Example of our compression scheme (MESSI) from A to Z}. Here $j=3$ and $k=2$, and we compress the embedding layer from figure~\ref{fig:fc}: (i) find the set of $k=2$ subspaces, each of dimension $j=3$, that minimizes the sum of squared distances from each point (row in $A$) to its \emph{closest} subspace. (ii) Partition the rows of $A$ into $k=2$ different subsets $A^1$ and $A^2$, where two rows are in the same subset if there closest subspace is the same, (iii) for each subset, factor its corresponding matrix into two smaller matrices based on its closest subspace to obtain the $2k=4$ matrices $U^1,V^1,U^2$ and $V^2$ (where for every $i\in \br{1,\cdots,k}$, the matrix $U^iV^i$ is a low ($j=3$) rank approximation for $A^i$), (iii) replace the original fully connected (embedding) layer by $2$ layers, where in the first (red color) we have $k=2$ parallel fully connected layers for (initialized by) $U^1$ and $U^2$ as in the figure, and the second (blue color) is a fully connected layer with all the previews $k=2$, and its weights corresponds to $V^1$ and $V^2$ as follow. For every $i\in \br{1,\cdots,k}$, the weights form the $j=3$ neurons (nodes) that are connected in the previous layer with $U^i$ are initialized by $V^i$.
    The result is a compressed layer that consists of $nj + kjd= 20\times3 + 2\times3\times10=120$ parameters. See more details in the figure.  }
    \label{fig:ksvd}
\end{figure}

We answer the above open questions by suggesting the following contributions.
In short, the answers are:
\begin{enumerate}[label=(\roman*)]
\item In all the real-world networks that we tested, it is almost always better to assume $k\geq2$ distributions rather than a single one that generated the data. It is better in the sense that the resulting accuracy of the network is better compared to $k=1$ (SVD) for the same compression rate.
\item While approximating the global minimum is Max-SNP-Hard, our experiments show that we can efficiently compute many local minima and take the smallest one. We then explain how to encode the result back into the network. This is by suggesting a new embedding layer architecture that we call MESSI (Multiple (parallel) Estimated SVDs for Smaller Intralayers); see Figure~\ref{fig:ksvd}. Extensive experimental results show significant improvement.
\end{enumerate}

\begin{figure}[h]
    \includegraphics[width=\textwidth]{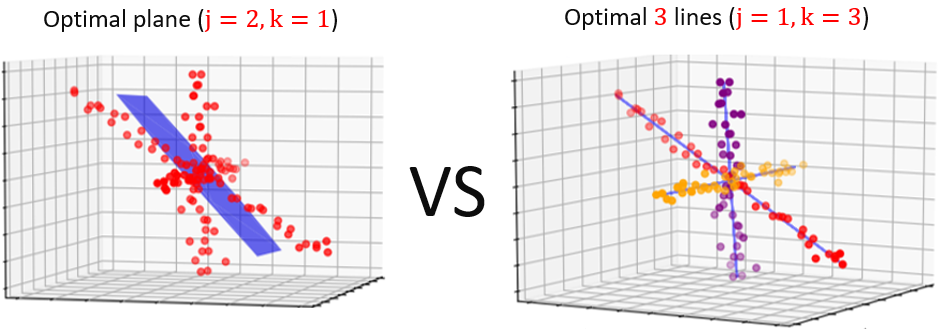}
    \caption{Why $k$ subspaces? Here, we have $n=120$ data points in $\REAL^3$ that are spread around $k=3$ lines ($j=1$). Factoring this data based on the optimal plane $P$ results with large errors, since some points are far from this plane as can be seen in the left hand side of the figure. On the right hand side, factoring the data based the $3$ optimal lines $\ell_1,\ell_2$, and $\ell_3$ gives a much smaller errors. Also, storing the factorization based on the plane $P$ requires $2(120 + 3) = 246$ parameters, compared to $120\times 1 + 3\times 3 = 129$ parameters based on $\ell_1,\ell_2$, and $\ell_3$. I.e., less memory and a better result.}
    \label{fig:planevslines}
\end{figure}

\paragraph{Computational Geometry meets Deep Learning. }
Our technique also constructs the matrix $A\in\REAL^{n\times d}$ from a given embedding layer. However, inspired by the geometric intuition from the previous section, we suggest to approximate the $n$ rows of $A$ by clustering them to $k\geq 2$ subspaces instead of one. More precisely, given an integer $k\geq1$ we aim to compute a set of $k$ subspaces in $\REAL^d$, each of dimension $j$, that will minimize the sum over every squared distance of every point (row in $A$) to its nearest subspace. This can be considered as a combination of $j$-rank or $j$-subspace approximation, as defined above, and $k$-means clustering. In the $k$-means clustering problem we wish to approximate $n$ points by $k$ center \emph{points} that minimizes the sum over squared distance between every point to its nearest center. In our case, the $k$ centers points are replaced by $k$ subspaces, each of dimension $j$. In computational geometry, this type of problem is called \emph{projective clustering}; see Figure~\ref{fig:planevslines}.

\paragraph{From Embedding layer to Embedding layers.} The result of the above technique is a set of $k$ matrices $A_j^1,\cdots,A_j^k$, each of rank $j$ and dimension $n_i\times d$ where the $i$th matrix corresponds to the cluster of $n_i$ points that were projected on the $i$th $j$-dimensional subspace. Each of those matrices can be factored into two smaller matrices (due to its low rank), i.e., for every $i\in \br{1,\cdots,k}$, we have $A^i_j =U^iV^i$, where $U^i \in \REAL^{n_i \times j}$, and $V^i\in \REAL^{j \times d}$. To plug these matrices as part of the final network instead of the embedded layer, we suggest to encode these matrices via $k$ parallel sub-layers as described in what follows and illustrated in Figure~\ref{fig:ksvd}.

\paragraph{Our pipeline: MESSI. }
We construct our new architecture as follows.
We use $A$ to refer to the $n \times d$ matrix from the embedding layer we seek to compress.  The input to our pipeline is the matrix $A$, positive integers $j$ and $k$, and (for the final step) parameters for the fine-tuning.

\begin{enumerate}
\item Treating the $n$ rows of $A$ as $n$ points in $\REAL^d$, compute an approximate $(k,j)$-projective clustering.  The result is $k$ subspaces in $\REAL^d$, each of dimension $j$, that minimize the sum of squared distances from each point (row in $A$) to its \emph{closest} subspace.
For the approximation, we compute a local minimum for this problem using the Expectation-Maximization (EM) method~\citep{dempster1977maximum}.

\item Partition the rows of $A$ into $k$ different subsets according to their nearest subspace from the previous step.  The result is submatrices $A^1, \ldots, A^k$ where $A^i$ is a $n_i \times d$ matrix and $n_1 + \ldots + n_k = n$.

\item For each matrix $A^i$ where $1 \le i \le k$, factor it to two smaller matrices $U^i$ (of dimensions $n_i \times j$) and $V^i$ (of dimensions $j \times d$) such that $U^iV^i$ is the rank-$j$ approximation of $A^i$.

\item In the full network, replace the original fully-connected embedding layer by $2$ layers. The first layer is a parallelization of $k$ separate fully-connected layers, where for every $i\in\br{1,\cdots,k}$ the $i$th parallel layer consists of the matrix $U^i$, i.e., it has $n_i$ input neurons and $j$ output neurons. Here, each row of $A$ is mapped appropriately.
The second layer is by combining the matrices ${V^1,\cdots V^k}$. Each of the $k$ output vectors from the previous layer $u_1, \ldots, u_k$ are combined as ${V^1u_1 + \ldots + V^ku_k}$; see Figure~\ref{fig:ksvd} for an illustration.

\item Fine-tune the network.
\end{enumerate}

The result is a compressed embedding layer.  Every matrix $U^i$ has $n_ij$ parameters, and the matrix $V^i$ has $jd$ parameters.  Therefore the compressed embedding layer consists of $nj + kjd$ parameters, in comparison to the uncompressed layer of $nd$ parameters.

\paragraph{Practical Solution. }The projective clustering problem is known to be Max-SNP-hard even for $d=2$ and $j=2$, for any approximation factor that is independent of $n$.
Instead, we suggest to use an algorithm that provably converges to a local minimum via the Expectation-Maximization (EM) method~\citep{dempster1977maximum}, which is a generalization of the well known Lloyd algorithm~\citep{lloyd1982least}.
The resulting clusters and factorizations are used to determine the new architecture and its initial weights; see Figure~\ref{fig:ksvd} for more details. We run on instances of AWS Amazon EC2 cloud, and detail our results in the next section.

\paragraph{Open code and networks. } Complete open code to reproduce the resulting networks is provided. We expect it to be useful for future research, and give the following few examples.

\subsection{Generalizations and Extensions.}\label{sec:Genaralization}
Our suggested architecture can be generalized and extended to support many other optimization functions that may be relevant for different types of datasets, tasks or applications besides NLP.
\paragraph{$\ell^q$-error. }For simplicity, our suggested approach aims to minimize sum of \emph{squared} distances to $k$ subspaces. However, it can be easily applied also to sum of distances from the points to the subspace. In this case, we aim to compute the maximum-likelihood of the generating subspaces assuming a Laplacian instead of Gaussian distribution. More generally, we may want to minimize the sum over every distance to the power of $q>0$., i.e., we take the $q$-norm $\norm{err}_q$ where $err$ is the distance between a point to its projection on its closest subspace.

Even for $k=1$ recent results of~\citet{tukan2020compressed} show improvement over SVD.

Observe that given the optimal subpaces, the system architecture in these cases remains the same as ours in Figure~\ref{fig:ksvd}, and a local minimum can still be obtained by the suggested algorithm. The only difference is that the SVD computation of the optimal subspace for a cluster of points ($k=1$) should be replaced by more involved approximation algorithm for computing the subspace that minimizes sum over distances to the power of $q$; see e.g.~\citet{tukan2020compressed,clarkson2015input}.

\textbf{Distance functions. }Similarly, we can replace the Euclidean $\ell_2$-distance by e.g. the Manhattan distance which is the $\ell_1$-norm between a point $x$ and its projection, i.e., $\norm{x-x'}_1$ or sum of differences between the corresponding entries, instead of sum of squared entries, as in the Euclidean distance $\norm{x-x'}_2$ in this paper. More generally, we may use the $\ell_p$ distance $\norm{x-x'}_p$, or even non-distance functions such as M-Estimators that can handle outliers (as in~\citet{tukan2020coresets}) by replacing $\dist(p,x)$ with $\min\br{\dist(p,x),t}$ where $t>0$ is constant (threshold) that makes sure that far away points will not affect the overall sum too much.

From an implementation perspective, the EM-algorithm for $k$-subspaces uses a $k=1$ solver routine as a blackbox.  Therefore extending to other distance functions is as simple as replacing the SVD solver (the $k=1$ for Euclidean distance) by the corresponding solver for $k=1$.

\textbf{Non-uniform dimensions. }In this paper we assume that $k$ subspaces approximate the input points, and each subspace has dimension exactly $j$, where $j,k\geq1$ are given integers. A better strategy is to allow each subspace to have a different dimension, $j_i$ for every $i\in\br{1,\cdots,k}$, or add a constraint only on the sum $j_1+\cdots+j_k$ of dimensions. Similarly, the number $k$ may be tuned as in our experimental results. Using this approach we can improve the accuracy and enjoy the same compression rate. This search or parameter tuning, however, might increase the computation time of the compressed network. It also implies layers of different sizes (for each subspace) in Figure~\ref{fig:ksvd}.

\textbf{Dictionary Learning. }Our approach of projective clustering is strongly related to Dictionary Learning~\citep{tosic2011dictionary,mairal2009supervised}. Here, the input is a matrix $A\in\REAL^{n\times d}$ and the output is a ``dictionary" $V^T\in\REAL^{d\times j}$ and projections or atoms which are the rows of $U\in\REAL^{n\times j}$ that minimize $\norm{A-UV}$ under some norm. It is easy to prove that $UV$ is simply the $j$-rank approximation of $A$, as explained in Section~\ref{sec:matrixfactorization}. However, if we have additional constraints,  such as that every row of $U$ should have, say, only $k=1$ non-zero entries, then geometrically the columns of $V^T$ are the $j$ lines that intersects the origin and minimize the sum of distances to the points. For $k>1$ every point is projected onto the subspace that minimizes its distance and is spanned by $k$ columns of $V^T$.

\textbf{Coresets. }Coresets are a useful tool, especially in projective clustering, to reduce the size of the input (compress it in some sense) while preserving the optimal solution or even the sum of distances to any set of $k$ subspaces. However, we are not aware of any efficient implementations and the dependency on $d$ and $k$ is usually exponential as in~\citet{edwards2005no}. A natural open problem is to compute more efficient and practical coresets for projective clustering.

\section{Experimental Results}
\label{sec:results}
\textbf{GLUE benchmark.} We run all of our experiments on the General Language Understanding Evaluation (GLUE) benchmark~\citep{wang2018glue}. It is widely-used collection of $9$ datasets for evaluating natural language understanding systems.

\textbf{Networks. }We compress the following two networks: (i) RoBERTa~\citep{liu2019RoBERTa}, it consists of $120$ millions parameters, and its embedding layer has $38.9$ million parameters ($32.5\%$ of the entire network size), and (ii) DistilBERT~\citep{sanhdistilbert} consists of $66$ million parameters, and its embedding layer has $23.5$ million parameters ($35.5\%$ of the entire network size).

\textbf{Software, and Hardware. }All the experiments were conducted on a AWS c$5$a.16xlarge machine with $64$ CPUs and $128$ RAM [GiB].
To build and train networks, we used the suggested implementation at the Transformers~\footnote{https://github.com/huggingface/transformers} library from HuggingFace~\citep{wolf2019huggingface} (Transformers version $3.1.0$, and PyTorch version $1.6.0$~\citep{paszke2017automatic}). 

\textbf{The setup. }
All our experiments are benchmarked against their publicly available implementations of the DistilBERT and RoBERTa models, fine-tuned for each task, which was in some cases higher and in other cases lower than the values printed in the publications introducing these models. Given an embedding layer from a network that is trained on a task from GLUE, an integer $k\geq1$, and an integer $j\geq1$. We build and initialize a new architecture that replaces the original embedding layer by two smaller layers as explained in Figure~\ref{fig:ksvd}. We then fine tune the resulted network for $2$ epochs. We ran the same experiments for several values of $k$ and $j$ that defines different compression rater. We compete with the standard matrix factorization approach in all experiments.

\begin{figure}[h]
    \includegraphics[width=0.33\textwidth]{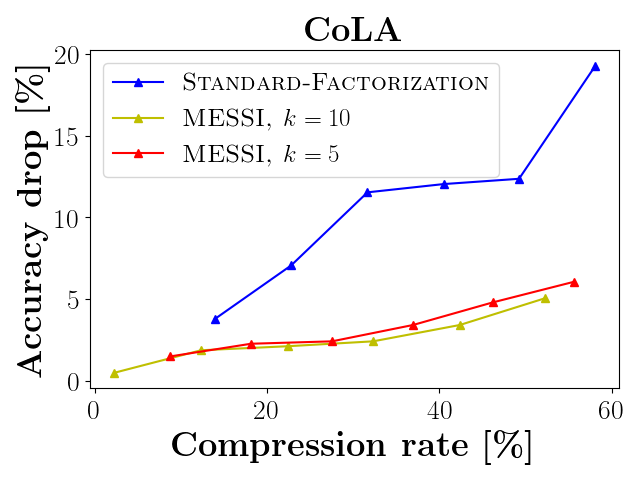}
    \includegraphics[width=0.33\textwidth]{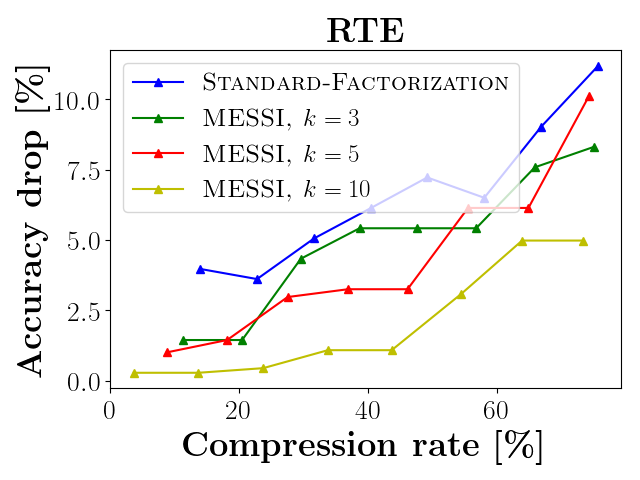}
    \includegraphics[width=0.33\textwidth]{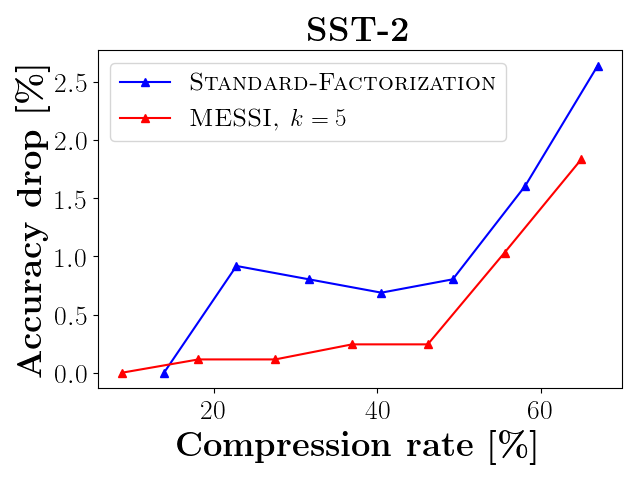}
    
    \includegraphics[width=0.33\textwidth]{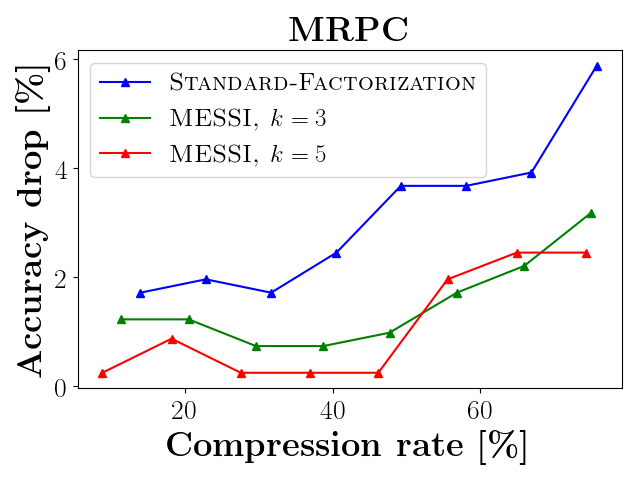}
    \includegraphics[width=0.33\textwidth]{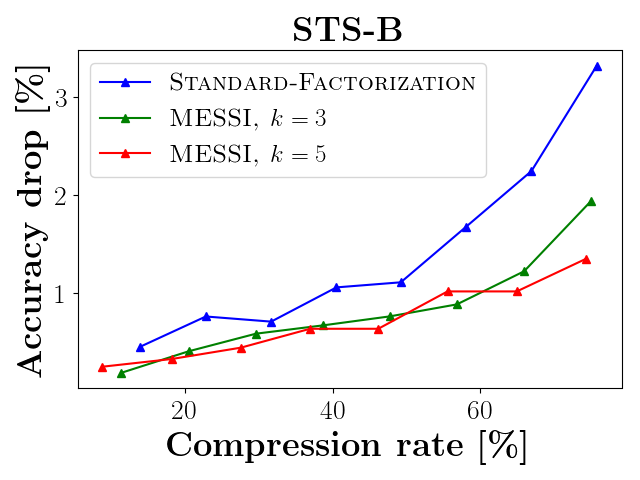}
    \includegraphics[width=0.33\textwidth]{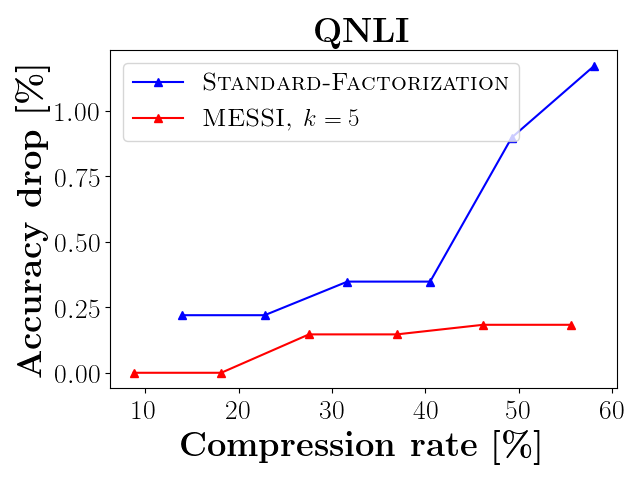}
\caption{Results on RoBERTa: Accuracy drop as a function of compression rate, with fine tuning for $2$ epochs after compression.  To illustrate the dependence of MESSI on the choice of $k$, we have plotted several contours for constant-$k$.  As the reader will notice, the same dataset may be ideally handled by different values of $k$ depending on the desired compression.}\label{fig:results1}
\end{figure}

\begin{figure}[h]
    \includegraphics[width=0.33\textwidth]{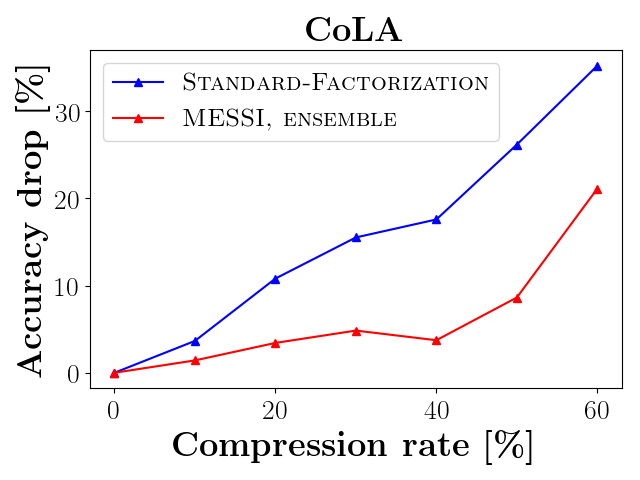}
    \includegraphics[width=0.33\textwidth]{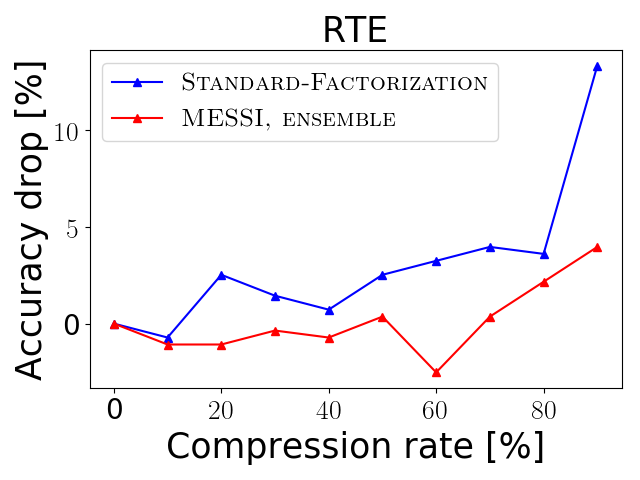}
    \includegraphics[width=0.33\textwidth]{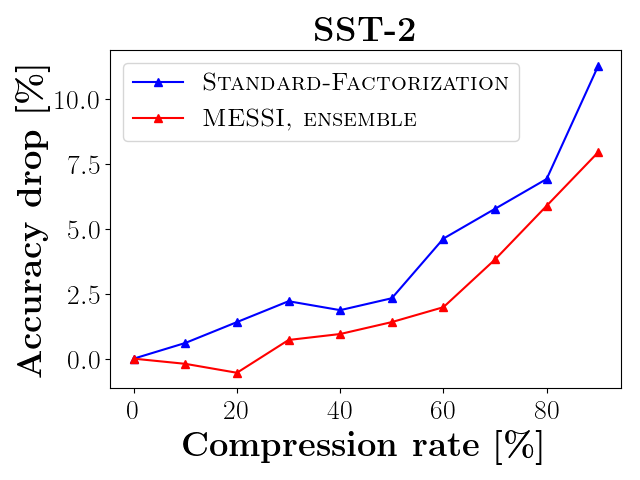}
    
    \includegraphics[width=0.24\textwidth]{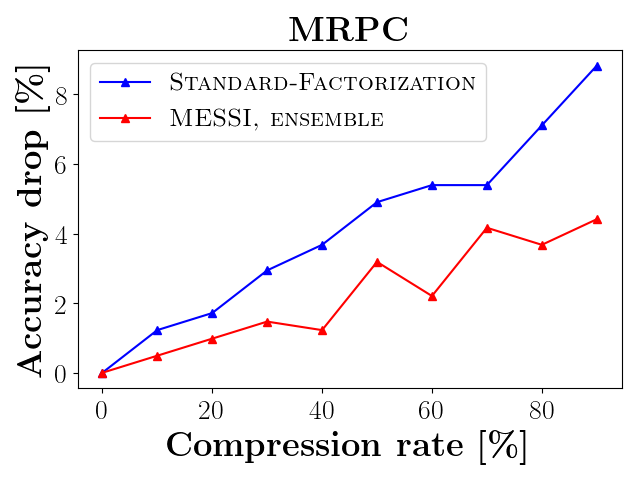}
    \includegraphics[width=0.24\textwidth]{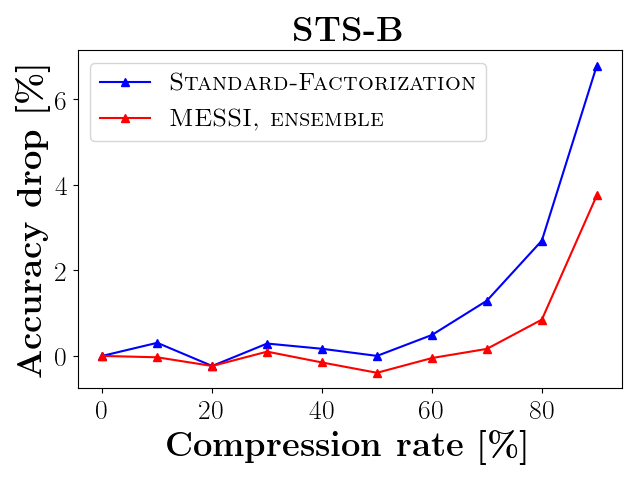}
    \includegraphics[width=0.24\textwidth]{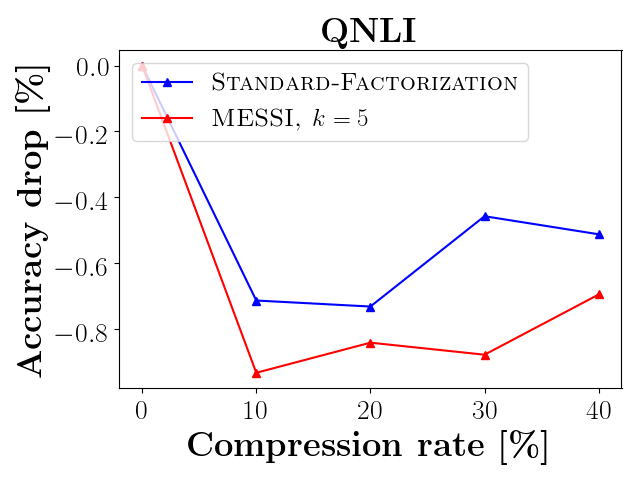}
    \includegraphics[width=0.24\textwidth]{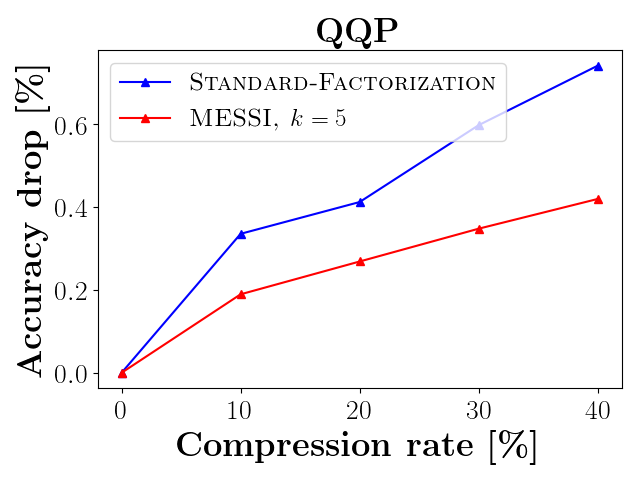}
\caption{Results on DistilBERT: Accuracy drop as a function of compression rate, with fine tuning for $2$ epochs after compression.  The red line (MESSI, ensemble) is obtained by training models at several $k$ values and then evaluating the model that achieves the best accuracy on the training set.}\label{fig:results2}
\end{figure}

\textbf{Reported results. } (i) In Figures~\ref{fig:results1} and~\ref{fig:results2} the $x$-axis is the compression rate of the embedding layer, i.e. a compression of $40\%$ means the layer is $60\%$ its original size.
The $y$-axis is the accuracy drop (relative error) with respect to the original accuracy of the network (with fine tuning for $2$ epochs). In  Figure~\ref{fig:results1}, each graph reports the results for a specific task from the GLUE benchmark on RoBERTa, while Figure~\ref{fig:results2} reports the results of DistilBERT.
(ii) On the task WNLI we achieved $0$ error on both networks using the two approaches of SVD and our approach until $60\%$ compression rate, so we did not add a figure on it.
(iii) In RoBERTa, we ran one compression rate on MNLI due to time constraints, and we achieved similar results in both techniques.
We compressed $45\%$ of the embedding layer, based on our technique with $k=5$ and $j=384$ to obtain only $0.61\%$ drop in accuracy with fine tuning and $4.2\%$ without, this is compared to $0.61\%$ and $13.9\%$ respectively for the same compression rate via SVD factorization.
In DistilBERT, we compressed $40\%$ of the embedding layer with $k=4$ and achieved a $0.1\%$ \textit{increase} in accuracy after fine-tuning, as compared to a $0.05\%$ drop via SVD factorization (on MNLI).
(iv) Finally, Figures~\ref{fig:results_nofietunning} in the appendix, shows the accuracy drop as a function of the compression rate without fine tuning.

\subsection{discussion, conclusion and future work}
As shown by Figures~\ref{fig:results1} and~\ref{fig:results2}, our approach outperforms the traditional SVD factorization. In all experiments, our suggested compression achieves better accuracy for the same compression rate compared to the traditional SVD. For example, in RobERTa, we compress $43\%$ of the embedding layer with less that $0.8\%$ average drop in accuracy, this is compared to the $3\%$ drop in the standard technique for a smaller compression rate of $40\%$. In DistilBERT, we achieved $40\%$ compression of the embedding layer while incurring only a $0.5\%$ average drop in accuracy over all nine GLUE tasks, compared to a $2.8\%$ drop using the existing SVD approach.
We observed that our technique shines mainly when the network is efficient, and any small change will lead to large error, e.g., as in the CoLA/RTE/MRPC graph of Figure~\ref{fig:results1}. Although we achieve better results in all of the cases, but here the difference is more significant (up to $10\%$), since our compressed layer approximates the original layer better than SVD, the errors are smaller, and the accuracy is better.
Finally, Figure~\ref{fig:results_nofietunning} shows clearly that even without fine tuning, the new approach yields more accurate networks. 
Hence, we can fine tune for smaller number of epochs and achieve higher accuracy and smaller networks.

\textbf{Future work includes:} (i) Experiments on other networks and data sets both from the field of NLP and outside it, (ii) an inserting experiment is to modify the ALBERT network~\citep{lan2019albert}, by changing its embedding layer architecture (that consists of two layers based on the standard matrix factorization) to the suggested architecture in this paper, while maintaining the same number of parameters, and to check if this modification improved its accuracy, and (iii) try the suggested generalizations and extensions from section~\ref{sec:Genaralization}, where we strongly believe they will allow us to achieve even better results.

\bibliographystyle{plainnat}
\bibliography{general}

\begin{thebibliography}{38}
\providecommand{\natexlab}[1]{#1}
\providecommand{\url}[1]{\texttt{#1}}
\expandafter\ifx\csname urlstyle\endcsname\relax
  \providecommand{\doi}[1]{doi: #1}\else
  \providecommand{\doi}{doi: \begingroup \urlstyle{rm}\Url}\fi

\bibitem[Acharya et~al.(2019)Acharya, Goel, Metallinou, and
  Dhillon]{acharya2019online}
Anish Acharya, Rahul Goel, Angeliki Metallinou, and Inderjit Dhillon.
\newblock Online embedding compression for text classification using low rank
  matrix factorization.
\newblock In \emph{Proceedings of the AAAI Conference on Artificial
  Intelligence}, volume~33, pages 6196--6203, 2019.

\bibitem[Clarkson and Woodruff(2015)]{clarkson2015input}
Kenneth~L Clarkson and David~P Woodruff.
\newblock Input sparsity and hardness for robust subspace approximation.
\newblock In \emph{2015 IEEE 56th Annual Symposium on Foundations of Computer
  Science}, pages 310--329. IEEE, 2015.

\bibitem[Dai and Le(2015)]{dai2015semi}
Andrew~M Dai and Quoc~V Le.
\newblock Semi-supervised sequence learning.
\newblock In \emph{Advances in neural information processing systems}, pages
  3079--3087, 2015.

\bibitem[Dempster et~al.(1977)Dempster, Laird, and Rubin]{dempster1977maximum}
Arthur~P Dempster, Nan~M Laird, and Donald~B Rubin.
\newblock Maximum likelihood from incomplete data via the em algorithm.
\newblock \emph{Journal of the Royal Statistical Society: Series B
  (Methodological)}, 39\penalty0 (1):\penalty0 1--22, 1977.

\bibitem[Devlin et~al.(2019)Devlin, Chang, Lee, and
  Toutanova]{devlin-etal-2019-bert}
Jacob Devlin, Ming-Wei Chang, Kenton Lee, and Kristina Toutanova.
\newblock {BERT}: Pre-training of deep bidirectional transformers for language
  understanding.
\newblock In \emph{Proceedings of the 2019 Conference of the North {A}merican
  Chapter of the Association for Computational Linguistics: Human Language
  Technologies, Volume 1 (Long and Short Papers)}, pages 4171--4186,
  Minneapolis, Minnesota, June 2019. Association for Computational Linguistics.
\newblock \doi{10.18653/v1/N19-1423}.
\newblock URL \url{https://www.aclweb.org/anthology/N19-1423}.

\bibitem[Edwards and Varadarajan(2005)]{edwards2005no}
Michael Edwards and Kasturi Varadarajan.
\newblock No coreset, no cry: Ii.
\newblock In \emph{International Conference on Foundations of Software
  Technology and Theoretical Computer Science}, pages 107--115. Springer, 2005.

\bibitem[Fan et~al.(2019)Fan, Grave, and Joulin]{fan2019reducing}
Angela Fan, Edouard Grave, and Armand Joulin.
\newblock Reducing transformer depth on demand with structured dropout.
\newblock In \emph{International Conference on Learning Representations}, 2019.

\bibitem[Gordon(2019)]{gordon_2019}
Mitchell~A. Gordon.
\newblock All the ways you can compress bert.
\newblock
  http://mitchgordon.me/machine/learning/2019/11/18/all-the-ways-to-compress-BERT.html,
  2019.

\bibitem[Gordon et~al.(2020)Gordon, Duh, and Andrews]{gordon2020compressing}
Mitchell~A Gordon, Kevin Duh, and Nicholas Andrews.
\newblock Compressing bert: Studying the effects of weight pruning on transfer
  learning.
\newblock \emph{arXiv preprint arXiv:2002.08307}, 2020.

\bibitem[Guo et~al.(2019)Guo, Liu, Mungall, Lin, and Wang]{guo2019reweighted}
Fu-Ming Guo, Sijia Liu, Finlay~S Mungall, Xue Lin, and Yanzhi Wang.
\newblock Reweighted proximal pruning for large-scale language representation.
\newblock \emph{arXiv preprint arXiv:1909.12486}, 2019.

\bibitem[Jiao et~al.(2019)Jiao, Yin, Shang, Jiang, Chen, Li, Wang, and
  Liu]{jiao2019tinybert}
Xiaoqi Jiao, Yichun Yin, Lifeng Shang, Xin Jiang, Xiao Chen, Linlin Li, Fang
  Wang, and Qun Liu.
\newblock Tinybert: Distilling bert for natural language understanding.
\newblock \emph{arXiv preprint arXiv:1909.10351}, 2019.

\bibitem[Lan et~al.(2019)Lan, Chen, Goodman, Gimpel, Sharma, and
  Soricut]{lan2019albert}
Zhenzhong Lan, Mingda Chen, Sebastian Goodman, Kevin Gimpel, Piyush Sharma, and
  Radu Soricut.
\newblock Albert: A lite bert for self-supervised learning of language
  representations.
\newblock In \emph{International Conference on Learning Representations}, 2019.

\bibitem[Le and Mikolov(2014)]{le2014distributed}
Quoc Le and Tomas Mikolov.
\newblock Distributed representations of sentences and documents.
\newblock In \emph{International conference on machine learning}, pages
  1188--1196, 2014.

\bibitem[Liu et~al.(2019{\natexlab{a}})Liu, Wang, Lin, Socher, and
  Xiong]{liu2019attentive}
Linqing Liu, Huan Wang, Jimmy Lin, Richard Socher, and Caiming Xiong.
\newblock Attentive student meets multi-task teacher: Improved knowledge
  distillation for pretrained models.
\newblock \emph{arXiv preprint arXiv:1911.03588}, 2019{\natexlab{a}}.

\bibitem[Liu et~al.(2019{\natexlab{b}})Liu, Ott, Goyal, Du, Joshi, Chen, Levy,
  Lewis, Zettlemoyer, and Stoyanov]{liu2019RoBERTa}
Yinhan Liu, Myle Ott, Naman Goyal, Jingfei Du, Mandar Joshi, Danqi Chen, Omer
  Levy, Mike Lewis, Luke Zettlemoyer, and Veselin Stoyanov.
\newblock Roberta: A robustly optimized bert pretraining approach.
\newblock \emph{arXiv preprint arXiv:1907.11692}, 2019{\natexlab{b}}.

\bibitem[Lloyd(1982)]{lloyd1982least}
Stuart Lloyd.
\newblock Least squares quantization in pcm.
\newblock \emph{IEEE transactions on information theory}, 28\penalty0
  (2):\penalty0 129--137, 1982.

\bibitem[Mairal et~al.(2009)Mairal, Ponce, Sapiro, Zisserman, and
  Bach]{mairal2009supervised}
Julien Mairal, Jean Ponce, Guillermo Sapiro, Andrew Zisserman, and Francis~R
  Bach.
\newblock Supervised dictionary learning.
\newblock In \emph{Advances in neural information processing systems}, pages
  1033--1040, 2009.

\bibitem[McCarley(2019)]{mccarley2019pruning}
J~Scott McCarley.
\newblock Pruning a bert-based question answering model.
\newblock \emph{arXiv preprint arXiv:1910.06360}, 2019.

\bibitem[Michel et~al.(2019)Michel, Levy, and Neubig]{michel2019sixteen}
Paul Michel, Omer Levy, and Graham Neubig.
\newblock Are sixteen heads really better than one?
\newblock In \emph{Advances in Neural Information Processing Systems}, pages
  14014--14024, 2019.

\bibitem[Mikolov et~al.(2013)Mikolov, Sutskever, Chen, Corrado, and
  Dean]{mikolov2013distributed}
Tomas Mikolov, Ilya Sutskever, Kai Chen, Greg~S Corrado, and Jeff Dean.
\newblock Distributed representations of words and phrases and their
  compositionality.
\newblock In \emph{Advances in neural information processing systems}, pages
  3111--3119, 2013.

\bibitem[Mukherjee and Awadallah(2019)]{mukherjee2019distilling}
Subhabrata Mukherjee and Ahmed~Hassan Awadallah.
\newblock Distilling transformers into simple neural networks with unlabeled
  transfer data.
\newblock \emph{arXiv preprint arXiv:1910.01769}, 2019.

\bibitem[Paszke et~al.(2017)Paszke, Gross, Chintala, Chanan, Yang, DeVito, Lin,
  Desmaison, Antiga, and Lerer]{paszke2017automatic}
Adam Paszke, Sam Gross, Soumith Chintala, Gregory Chanan, Edward Yang, Zachary
  DeVito, Zeming Lin, Alban Desmaison, Luca Antiga, and Adam Lerer.
\newblock Automatic differentiation in pytorch.
\newblock In \emph{NIPS-W}, 2017.

\bibitem[Peters et~al.(2018)Peters, Neumann, Iyyer, Gardner, Clark, Lee, and
  Zettlemoyer]{peters2018deep}
Matthew Peters, Mark Neumann, Mohit Iyyer, Matt Gardner, Christopher Clark,
  Kenton Lee, and Luke Zettlemoyer.
\newblock Deep contextualized word representations.
\newblock In \emph{Proceedings of the 2018 Conference of the North American
  Chapter of the Association for Computational Linguistics: Human Language
  Technologies, Volume 1 (Long Papers)}, pages 2227--2237, 2018.

\bibitem[Radford et~al.(2018)Radford, Narasimhan, Salimans, and
  Sutskever]{radford2018improving}
Alec Radford, Karthik Narasimhan, Tim Salimans, and Ilya Sutskever.
\newblock Improving language understanding by generative pre-training, 2018.

\bibitem[Radford et~al.(2019)Radford, Wu, Child, Luan, Amodei, and
  Sutskever]{radford2019language}
Alec Radford, Jeffrey Wu, Rewon Child, David Luan, Dario Amodei, and Ilya
  Sutskever.
\newblock Language models are unsupervised multitask learners.
\newblock \emph{OpenAI Blog}, 1\penalty0 (8):\penalty0 9, 2019.

\bibitem[Sanh et~al.(2019)Sanh, Debut, Chaumond, and Wolf]{sanhdistilbert}
Victor Sanh, Lysandre Debut, Julien Chaumond, and Thomas Wolf.
\newblock Distilbert, a distilled version of bert: smaller, faster, cheaper and
  lighter.
\newblock \emph{arXiv preprint arXiv:1910.01108}, 2019.

\bibitem[Shen et~al.(2020)Shen, Dong, Ye, Ma, Yao, Gholami, Mahoney, and
  Keutzer]{shen2020q}
Sheng Shen, Zhen Dong, Jiayu Ye, Linjian Ma, Zhewei Yao, Amir Gholami,
  Michael~W Mahoney, and Kurt Keutzer.
\newblock Q-bert: Hessian based ultra low precision quantization of bert.
\newblock In \emph{AAAI}, pages 8815--8821, 2020.

\bibitem[Sun et~al.(2019)Sun, Cheng, Gan, and Liu]{sun2019patient}
Siqi Sun, Yu~Cheng, Zhe Gan, and Jingjing Liu.
\newblock Patient knowledge distillation for bert model compression.
\newblock In \emph{Proceedings of the 2019 Conference on Empirical Methods in
  Natural Language Processing and the 9th International Joint Conference on
  Natural Language Processing (EMNLP-IJCNLP)}, pages 4314--4323, 2019.

\bibitem[Tang et~al.(2019)Tang, Lu, Liu, Mou, Vechtomova, and
  Lin]{tang2019distilling}
Raphael Tang, Yao Lu, Linqing Liu, Lili Mou, Olga Vechtomova, and Jimmy Lin.
\newblock Distilling task-specific knowledge from bert into simple neural
  networks.
\newblock \emph{arXiv preprint arXiv:1903.12136}, 2019.

\bibitem[Tosic and Frossard(2011)]{tosic2011dictionary}
Ivana Tosic and Pascal Frossard.
\newblock Dictionary learning.
\newblock \emph{IEEE Signal Processing Magazine}, 28\penalty0 (2):\penalty0
  27--38, 2011.

\bibitem[Tukan et~al.(2020{\natexlab{a}})Tukan, Maalouf, and
  Feldman]{tukan2020coresets}
Murad Tukan, Alaa Maalouf, and Dan Feldman.
\newblock Coresets for near-convex functions.
\newblock \emph{arXiv preprint arXiv:2006.05482}, 2020{\natexlab{a}}.

\bibitem[Tukan et~al.(2020{\natexlab{b}})Tukan, Maalouf, Weksler, and
  Feldman]{tukan2020compressed}
Murad Tukan, Alaa Maalouf, Matan Weksler, and Dan Feldman.
\newblock Compressed deep networks: Goodbye svd, hello robust low-rank
  approximation.
\newblock \emph{arXiv preprint arXiv:2009.05647}, 2020{\natexlab{b}}.

\bibitem[Wang et~al.(2018)Wang, Singh, Michael, Hill, Levy, and
  Bowman]{wang2018glue}
Alex Wang, Amanpreet Singh, Julian Michael, Felix Hill, Omer Levy, and Samuel
  Bowman.
\newblock Glue: A multi-task benchmark and analysis platform for natural
  language understanding.
\newblock In \emph{Proceedings of the 2018 EMNLP Workshop BlackboxNLP:
  Analyzing and Interpreting Neural Networks for NLP}, pages 353--355, 2018.

\bibitem[Wang et~al.(2019)Wang, Wohlwend, and Lei]{wang2019structured}
Ziheng Wang, Jeremy Wohlwend, and Tao Lei.
\newblock Structured pruning of large language models.
\newblock \emph{arXiv preprint arXiv:1910.04732}, 2019.

\bibitem[Wolf et~al.(2019)Wolf, Debut, Sanh, Chaumond, Delangue, Moi, Cistac,
  Rault, Louf, Funtowicz, et~al.]{wolf2019huggingface}
Thomas Wolf, Lysandre Debut, Victor Sanh, Julien Chaumond, Clement Delangue,
  Anthony Moi, Pierric Cistac, Tim Rault, R{\'e}mi Louf, Morgan Funtowicz,
  et~al.
\newblock Huggingface's transformers: State-of-the-art natural language
  processing.
\newblock \emph{ArXiv}, pages arXiv--1910, 2019.

\bibitem[Yang et~al.(2019)Yang, Dai, Yang, Carbonell, Salakhutdinov, and
  Le]{yang2019XLNet}
Zhilin Yang, Zihang Dai, Yiming Yang, Jaime Carbonell, Russ~R Salakhutdinov,
  and Quoc~V Le.
\newblock Xlnet: Generalized autoregressive pretraining for language
  understanding.
\newblock In \emph{Advances in neural information processing systems}, pages
  5753--5763, 2019.

\bibitem[Yu et~al.(2017)Yu, Liu, Wang, and Tao]{yu2017compressing}
Xiyu Yu, Tongliang Liu, Xinchao Wang, and Dacheng Tao.
\newblock On compressing deep models by low rank and sparse decomposition.
\newblock In \emph{Proceedings of the IEEE Conference on Computer Vision and
  Pattern Recognition}, pages 7370--7379, 2017.

\bibitem[Zafrir et~al.(2019)Zafrir, Boudoukh, Izsak, and
  Wasserblat]{zafrirq8bert}
Ofir Zafrir, Guy Boudoukh, Peter Izsak, and Moshe Wasserblat.
\newblock Q8bert: Quantized 8bit bert.
\newblock \emph{arXiv preprint arXiv:1910.06188}, 2019.

\end{thebibliography}
\clearpage
\appendix

\section{Implementation in Pytorch}

Since we did not find straight support for the new suggested architecture, we implemented it as follows. To represent the matrices ${V^1,\cdots,V^k}$ that are described is Section~\ref{sec:ourcont}, we concatenate them all to a one large matrix $V=[ (V^1)^T ,\cdots, (V^k)^T ]^T$ of $kj$ rows and $d$ columns, and we build a fully-connected layer the corresponds to $V$.
For the $k$ parallel layers (matrices) ${U^1, \cdots, U^k}$, we build one large sparse matrix $U$ of $n$ rows and $kj$ columns. Every row of this matrix has at least $(k-1)j$ zero entries, and at most $j$ non zero entries, where the non-zero entries of the $i$th row corresponds to the rows in matrix $V$ which encode the closest subspace to that row's point. 

Finally, during the fine tuning or training, we set those zero entries in $U$ as non-trainable parameters, and we make sure that after every batch of back-propagation they remain zero. Hence we have at most $nj$ non-zero entries (trainable parameters) in $U$ and $nj+ndk$ in total.

We hope that in the future, the suggested architecture will be implemented in the known Deep-Learning libraries so it can be easily used while taking advantage of the substantial time and space benefits presented in this paper.
\begin{figure}[h]
    \includegraphics[scale = 0.6]{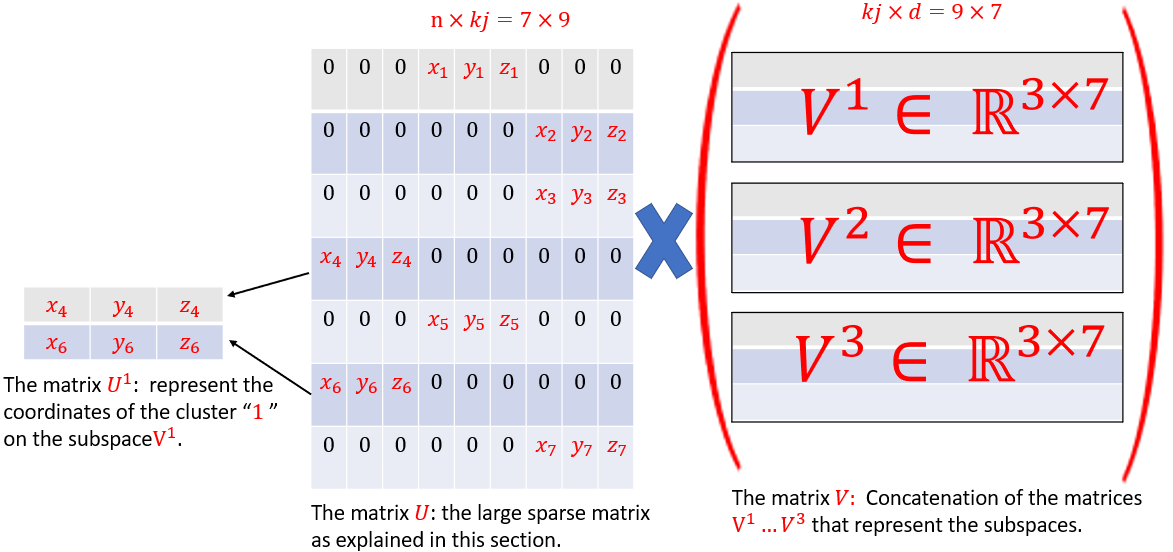}
    \caption{\textbf{Implementation.} Example of the factorization $A = UV$ in our implementation.  Here $n=7$ and $d=7$.  The matrix $U$ is built such that row $z$ contains a row from $U^i$ where point $z$ was partitioned to the $i^\text{th}$ subspace.  In this example, the $4^\text{th}$ and  $6^\text{th}$ rows were both clustered to the first subspace. Hence, the first 3 coordinates of the corresponding rows in the representation matrix $U$ are nonzero, and the other entries are zero.  In this way, we used $jk$ dimensions so that none of the $k$ subspaces of dimension $j$ interact.}
    \label{fig:code}
\end{figure}
\clearpage
\section{Results Before Fine Tuning}
\begin{figure}[h]

    \includegraphics[width=0.33\textwidth]{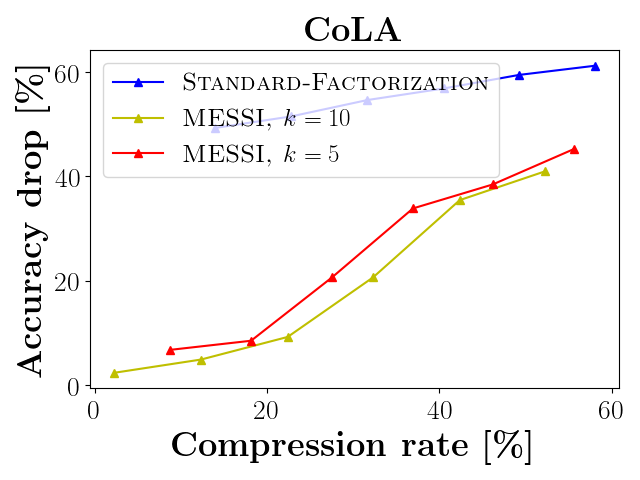}
    \includegraphics[width=0.33\textwidth]{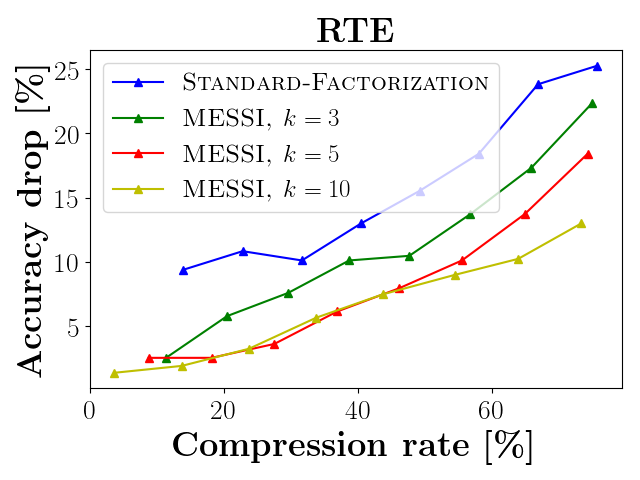}
    \includegraphics[width=0.33\textwidth]{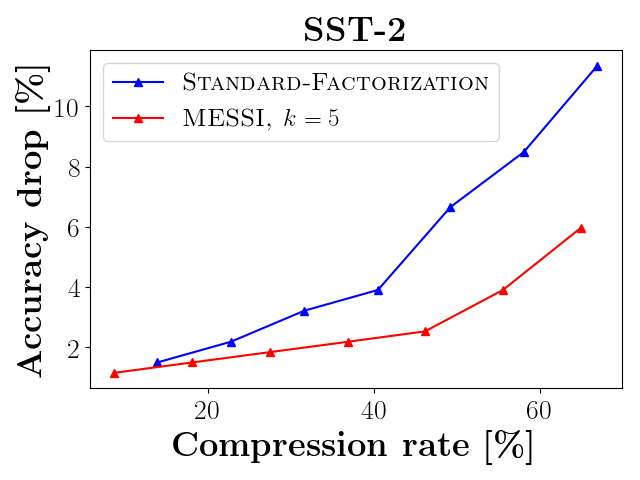}
    
    \includegraphics[width=0.33\textwidth]{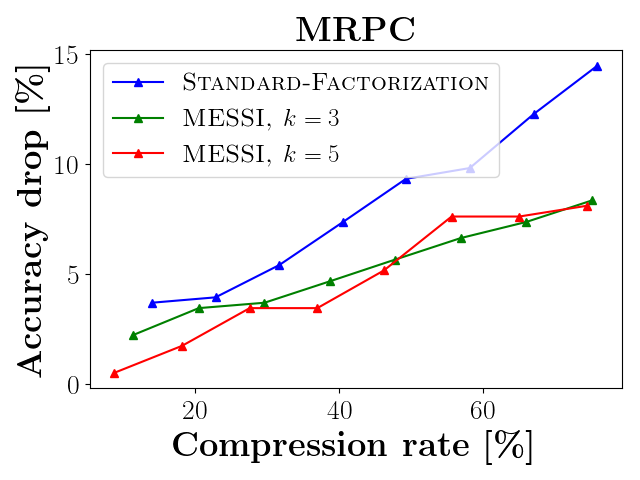}
    \includegraphics[width=0.33\textwidth]{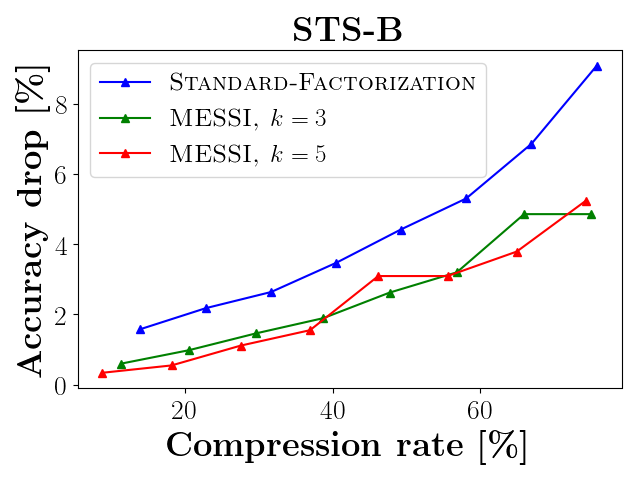}
    \includegraphics[width=0.33\textwidth]{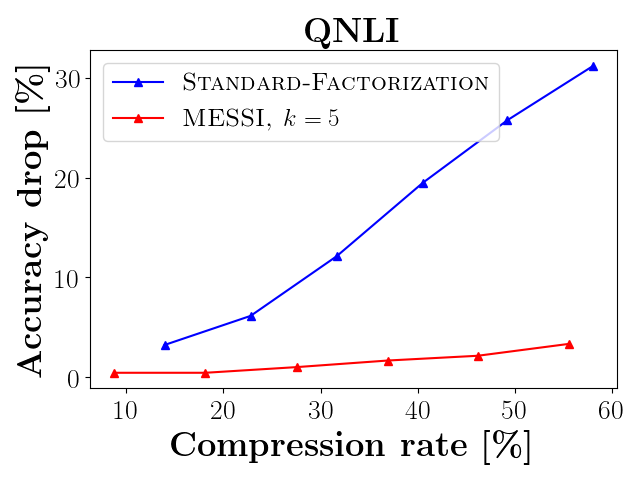}
\caption{Compressing RoBERTa results: Accuracy drop as a function of compression rate, without fine tuning.}\label{fig:results_nofietunning}
\end{figure}

\end{document}